\documentclass[
]{ceurart}

\usepackage{listings}
\usepackage{pifont}

\usepackage{url}

\usepackage{cleveref}
\Crefname{table}{Table}{Tables}
\Crefname{lstlisting}{Listing}{Listings}
\Crefname{algocf}{Algorithm}{Algorithms}

\usepackage{glossaries}

\newacronym{ai}{AI}{Artificial Intelligence}
\newacronym{dsg}{DSG}{Dynamic Scene Graph}
\newacronym{kg}{KG}{Knowledge Graph}
\newacronym{llm}{LLM}{Large Language Model}
\newacronym{rag}{RAG}{Retrieval Augmented Generation}
\newacronym{vlm}{VLM}{Vision Language Model}

\newcommand{\todo}[1]{\textcolor{red}{TODO: #1}}

\usepackage{listings}
\usepackage{xcolor}

\usepackage{tabularx}

\definecolor{codegreen}{rgb}{0,0.6,0}
\definecolor{codegray}{rgb}{0.5,0.5,0.5}
\definecolor{codepurple}{rgb}{0.58,0,0.82}
\definecolor{backcolour}{rgb}{0.95,0.95,0.92}

\begin{document}

\copyrightyear{2025}
\copyrightclause{Copyright for this paper by its authors.
  Use permitted under Creative Commons License Attribution 4.0
  International (CC BY 4.0).}

\conference{ESWC'25: Workshop on
LLM-Integrated Knowledge Graph Generation from Text (TEXT2KG), 
June 01--05, 2025, Portoroz, Slovenia}

\title{A Grounded Memory System For Smart Personal Assistants}



\author[1]{Felix Ocker}[%
email=felix.ocker@honda-ri.de,
]
\cormark[1]
\address[1]{Honda Research Institute Europe, Carl-Legien-Str. 30, 63073 Offenbach am Main, Germany}

\author[1]{Jörg Deigmöller}[%
]

\author[1]{Pavel Smirnov}[%
]

\author[1]{Julian Eggert}[%
]

\cortext[1]{Corresponding author.}



\begin{abstract}
A wide variety of agentic AI applications -- ranging from cognitive assistants for dementia patients to robotics -- demand a robust memory system grounded in reality.
In this paper, we propose such a memory system consisting of three components.
First, we combine Vision Language Models for image captioning and entity disambiguation with Large Language Models for consistent information extraction during perception.
Second, the extracted information is represented in a memory consisting of a knowledge graph enhanced by vector embeddings to efficiently manage relational information.
Third, we combine semantic search and graph query generation for question answering via Retrieval Augmented Generation.
We illustrate the system's working and potential using a real-world example.
\end{abstract}

\begin{keywords}
  Memory System \sep
  Ontology Construction \sep
  Retrieval Augmented Generation \sep
  GraphRAG \sep
  Grounding
\end{keywords}

\maketitle

\section{Introduction}
\label{sec:intro}

The emergence of \glspl{llm} has advanced conversational assistants beyond rule-based systems, enabling them to operate within a user’s perceptual and conceptual context.
For this, assistants must integrate stored knowledge with ongoing interactions to ensure that responses remain relevant and grounded.
\gls{rag} techniques combine \glspl{llm} with external knowledge bases and multimodal \glspl{llm} process diverse inputs, enabling richer, more context-aware interactions.
However, the need for assistants with personal and situational support based on a large-scale memory also highlights critical challenges.
First, to effectively deal with memories in the form of multimodal inputs, a robust conceptual framework is needed that acknowledges the role of space and time as fundamental dimensions of experience and memory.
Inspired by Kantian notions, which describe space and time as fundamental structuring elements imposed by the mind \cite{kant1908critique}, we recognize that memory systems must integrate these dimensions to maintain coherent, grounded knowledge.
Second, standard \gls{rag} stores information as disconnected snippets, failing to capture the relational dependencies needed for complex queries \cite{kashmira2024graph}.
Third, true situational awareness requires structured, concept-based retrieval and inference for more advanced reasoning and decision-making.
To address these challenges, we propose a novel approach for grounded memory-based personal assistants.
Our approach builds on a structured memory akin to human episodic and biographic memory, ensuring that information is pre-structured before inference rather than relying on on-the-fly conceptualization like standard \gls{rag}.
Each of these components addresses a specific challenge, resulting in three pillars:

\begin{enumerate}
    \item \textit{Grounded Perception}: Structure multimodal inputs with spatial and temporal awareness, categorizing them into actions, agents, and objects.
    \item \textit{Memory Graph}: Overcome standard \gls{rag} limitations by using a richer knowledge representation in the form of an ontological framework for representing memories, i.e., structuring interconnected concepts and enhancing memory versatility through semantic embeddings.
    \item \textit{Agentic Retrieval}: Use graph querying and expansion together with semantic search for improving coherence and context-awareness for complex queries.
\end{enumerate}

By combining these elements, our system enables assistants to deliver personalized, context-aware support with enhanced reasoning and decision-making.

\section{Related Work}


This section reviews related work for the three pillars of our memory system: grounded perception, memory graphs, and agentic retrieval.
Grounded perception organizes multimodal data into actions, agents, and objects with temporal awareness, forming structured action patterns \cite{eggert2020action}.
Memory graphs overcome standard \gls{rag} limitations by structuring knowledge to capture even implicit relations.
Agentic retrieval enhances reasoning via graph-based inference instead of relying solely on embedding similarity.

\subsection{Grounded Perception}


Multimodal perception has advanced with \glspl{llm}, aiding AI applications like robotics and surveillance. 
For instance, robots critically depend on their visual understanding capabilities for navigation \cite{al2024review} and object localization tasks \cite{armeni20193d,kim20193,rosinol20023d}. Recent work such as 3D \glspl{dsg} \cite{kim20193} and TASKOGRAPHY \cite{agia2022taskography} rely on creating structured models of the environment. However, perception in robotics usually does not focus on building a lifelong memory, but rather on creating a faithful representation of the current environment which could be recalled for specific tasks. For Embodied \gls{rag} \cite{xie2024embodied}, the authors build a structured semantic forest based on spatial proximity which can be used in combination with \glspl{llm} to support robotic navigation. 
Other examples of multimodal perception systems specializing in human activity recognition are systems for understanding long videos \cite{wang2023lifelongmemory}. Such systems, e.g., VideoAgent \cite{fan2024videoagent} and AMEGO \cite{goletto2024amego}, focus on person-object annotations, primarily tracking hand-object interactions without explicit action labeling. 
Effective memory-based assistants need persistent representations of actions, agents, and objects with contextual tracking.
Many multimodal perception systems offer contextualized understanding but lack structured long-term recall. Our approach integrates LLM-based perception with a structured graph-based memory to ensure interpretability and retrieval.
With advances in \glspl{llm}, many specialized environment recognition and action detection approaches are being replaced by multimodal \glspl{llm} \cite{li2025visual}.
In the context of this paper, we rely on multimodal \glspl{llm}, specifically \glspl{vlm}, for these tasks, since they generally provide more contextualized information for building a grounded memory system.
While \glspl{vlm} provide a flexible and context-aware understanding, they lack the structured memory needed for long-term, explainable recall by themselves.
Our approach addresses this by integrating \gls{vlm}-based perception into a structured, graph-based memory, ensuring that memories remain interpretable and retrievable.

\subsection{Memory Graphs}

A scalable memory is essential for assistants with personal support capabilities. 
Due to their benefits regarding the integration of heterogeneous data \cite{hogan2021knowledge}, knowledge graphs provide an excellent technological basis for such a memory \cite{eggert2025graph}.
For instance, TobuGraph \cite{kashmira2024graph} is an approach to transform pictures and conversations with a text-based chatbot into a memory graph. The authors demonstrate the limitations of the standard \gls{rag} approach for describing personal memories. In \cite{maniar2025mempal}, the authors describe MemPal - a wearable video-based conversation device for assisting elderly with memory impairments.
MemPal focuses on a use-case of finding lost objects and is evaluated on the effects of voice-enabled multimodal \glspl{llm}.
These systems address two deficiencies of standard \gls{rag} approaches: 1) The problem of scaling them to large-scale multimodal real-world scenarios and 2) the deficiencies in terms of representing complex memories of interconnected world entities. To address the second deficiency, the authors of \cite{li2024omniquery} describe a framework for capturing lifelong personal memories from images and videos by memorizing them via a natural language interface. The approach includes extracting a taxonomy of contextual information out of textual information obtained from videos and images, with contexts being described by time, location, people, visual elements of environment, activities and emotions. The extracted taxonomy is used for a special retrieval which augments semantic search. While this demonstrates that \gls{rag}-based approaches can be used to retrieve snippets of personal experiences, it lacks the power of relational memories as provided by underlying memory graphs. In this paper, we rely on a combination of \gls{rag} techniques with knowledge graphs for improving the retrieval capabilities of such systems.



\subsection{Agentic Retrieval}
\label{subsec:agentic-retrieval}

GraphRAG is a retrieval-augmented generation that enriches conventional \gls{rag} pipelines with a graph-based representation of knowledge. In standard \gls{rag} systems, semantic search is used to retrieve relevant text snippets from a vector store, which are provided as context to an \gls{llm} for question answering.
However, this chunk-oriented retrieval can miss deeper relationships and dependencies among pieces of information.
GraphRAG addresses this limitation by building and utilizing knowledge structured in graphs, enabling more coherent reasoning over interconnected facts.
One way of realizing GraphRAG is to conduct semantic search to find entry points in the graph and then expand the context for further relevant, but less explicit, information.
There are several suitable algorithms for graph expansion, PageRank being one of them \cite{gutierrez2024hipporag}.
Another approach to GraphRAG is to translate natural language queries into graph queries, such as Cypher, for structured database access \cite{ozsoy2024text2cypher}. 
Another GraphRAG application is presented in \cite{edge2024local}, where a knowledge graph is built from textual data. Instead of retrieving isolated text snippets, the system retrieves relational subgraphs relevant to a user query, which are then passed to an \gls{llm}.
By leveraging both textual and structured graph-based knowledge, this approach enables deeper reasoning over complex, interconnected facts, making it highly effective for answering intricate queries.
By leveraging a combination of these techniques, our system ensures more explainable and context-aware responses, combining the flexibility of text-based search with the expressiveness of a graph-based memory.

\section{Grounded Memory System Architecture}

\begin{figure}[htb]
  \centering
  \includegraphics[width=\linewidth]{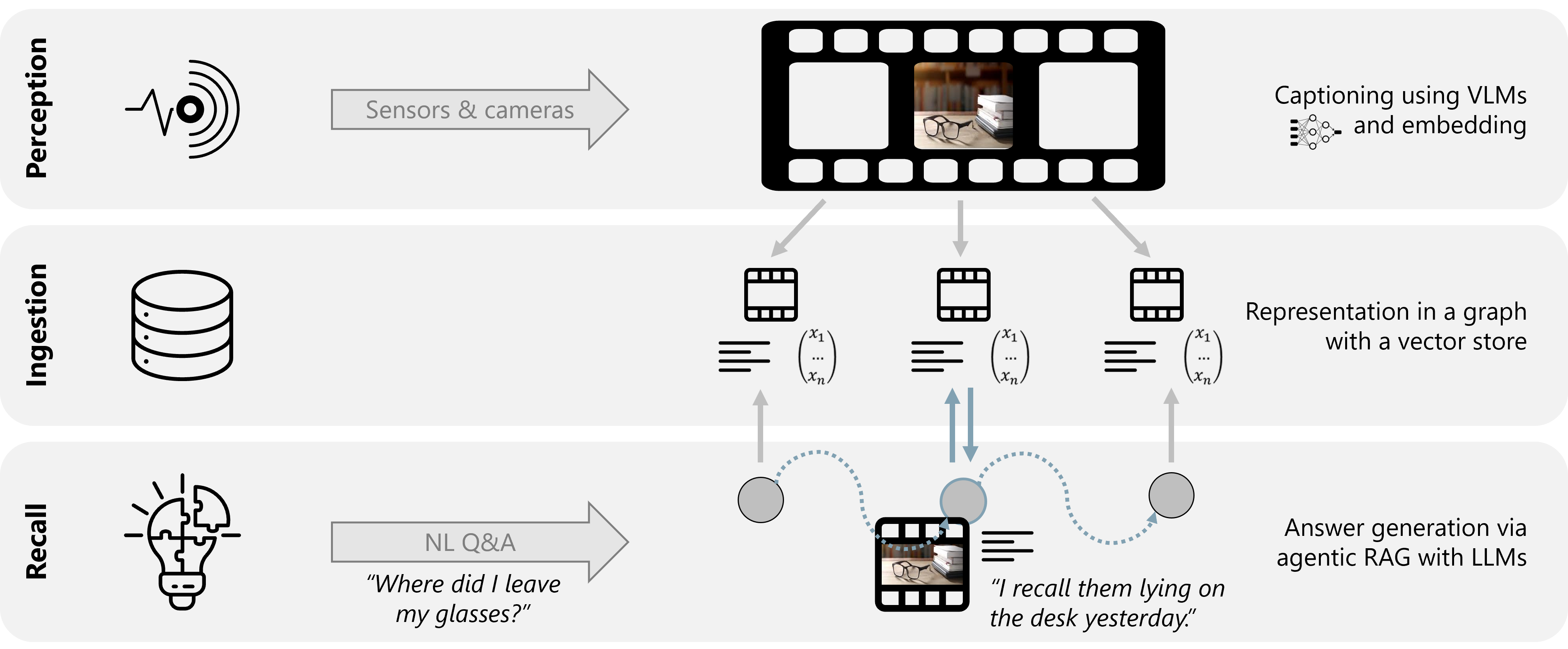}
  \caption{Memory system architecture overview.}
  \label{fig:phasesoverview}
\end{figure}

The memory system is based on a \textbf{schema} that revolves around textual notes, which are represented as nodes in a graph, cp.~\Cref{subsec:schema}.
Leveraging this schema, the memory system is designed to seamlessly capture, structure, and retrieve real-world observations through a three-phase process, see \Cref{fig:phasesoverview}.

In the \textbf{perception} phase, cp.~\Cref{subsec:perception}, cameras observe the environment, allowing a \gls{vlm} to generate descriptive captions for detected events.
While this can be extended for further modalities such as audio, we focus on visual inputs in the context of this paper.
During the \textbf{ingestion} phase, cp.~\Cref{subsec:population}, these images and captions undergo a structured analysis before being stored in a persistent knowledge graph. Unlike unstructured memory systems, this graph-based representation explicitly encodes who performed which action on which object, when, and where.
Finally, in the \textbf{recall} phase, cp.~\Cref{subsec:retrieval}, the system retrieves stored information for question answering, event verification, and intelligent recommendations.
Throughout the following, we rely on a video showing an individual in a home setting as a running example to exemplify the concepts introduced.


\subsection{Representing Memory Notes}
\label{subsec:schema}

\begin{figure}[ht]
  \centering
  \includegraphics[width=.8\linewidth]{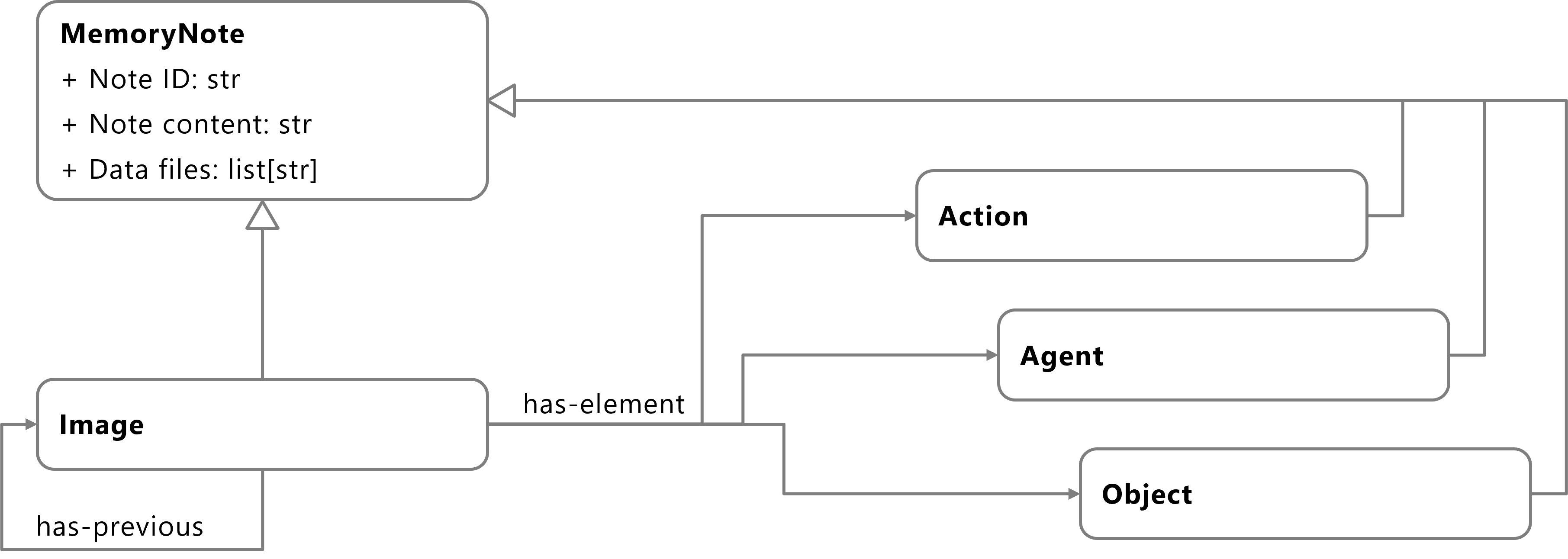}
  \caption{Schema for representing images and action patterns.}
  \label{fig:schema}
\end{figure}

The knowledge base builds on a schema for representing so-called memory notes, cp.~\Cref{fig:schema}.
A \textit{MemoryNote} can be used to describe a time period of arbitrary length and it can be generated from arbitrary sources, e.g., manually crafted for diary entries or generated automatically to describe a single frame in a video.
Each memory note is characterized by its note content, which is a natural language string, and an optional list of data files from which it has been created.
To create a structured representation, every memory note is also represented as a node in a knowledge graph.
For our application, we introduce \textit{Image} nodes, which are specialized memory notes that have an image caption as note content and that refer to an image as a data file. 
To create a structured representation, each \textit{MemoryNote} is linked to the entities it mentions, categorized as \textit{Agents} (\textit{"Who performed the action?"}), \textit{Objects} (\textit{"What was acted upon?"}), and \textit{Actions} (\textit{"What was done?"}).
Images are temporally ordered using \textit{has-previous} links,~cp.~\Cref{fig:schema}, and agents, objects, and actions are connected to the images they occur in via \textit{has-element} links.

\subsection{Perception}
\label{subsec:perception}

The perception phase captures raw video input images, and generates descriptive captions using \texttt{gpt-4o}'s \cite{achiam2024gpt4} vision capabilities, thus laying the foundation for a structured representation of events.

\begin{figure}[ht]
\centering
\includegraphics[width=\linewidth]{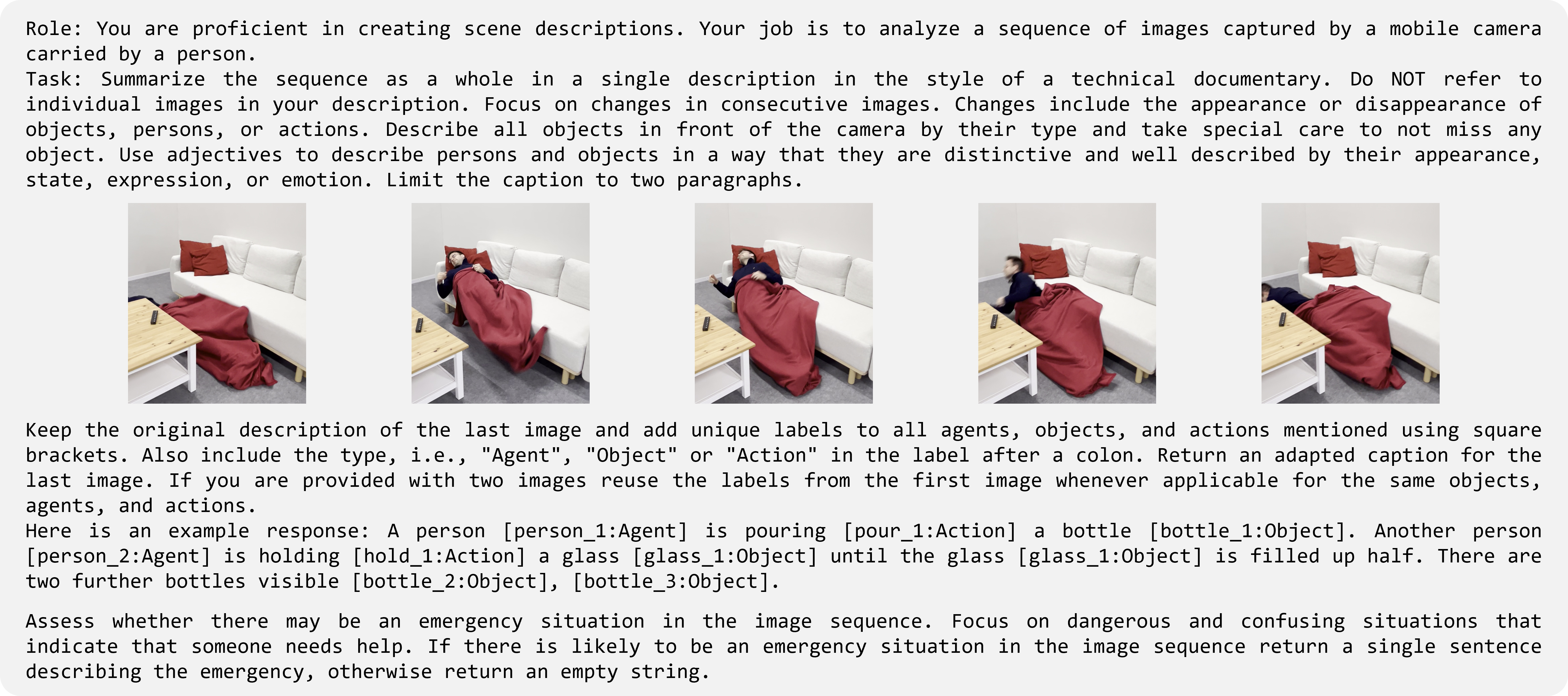}
\caption{Three-step \gls{vlm} prompt applied to an image sequence. Each processing window consists of multiple consecutive frames, analyzed together. The first and last frame overlap with adjacent windows to ensure continuity. Captions for the sequence are appended to the last frame.}
\label{fig:vlm}
\end{figure}

Unlike traditional video memory systems that passively store raw visual inputs, our system actively identifies and links key entities in the environment.
This includes detecting agents, objects, and their spatial relationships, forming a structured representation.
To achieve this, the system processes sequences of $n$ consecutive frames from the video stream using the prompt shown in \Cref{fig:vlm}. Each sequence is analyzed as a single unit, where the first and last frame overlap with adjacent sequences, ensuring temporal continuity. To find a balance between efficiency and accuracy, we caption only each n-th frame.
This strategy maintains temporal coherence, reduces redundant descriptions, and results in more accurate, context-aware scene summaries. Each described instance in the captions is indexed with a unique label in the format [\textit{label\_x}:\textit{Type}] to ensure consistent tracking across frames.

\subsection{Knowledge Graph and Vector Store Population}
\label{subsec:population}

\begin{figure}[ht]
  \centering
  \includegraphics[width=\linewidth]{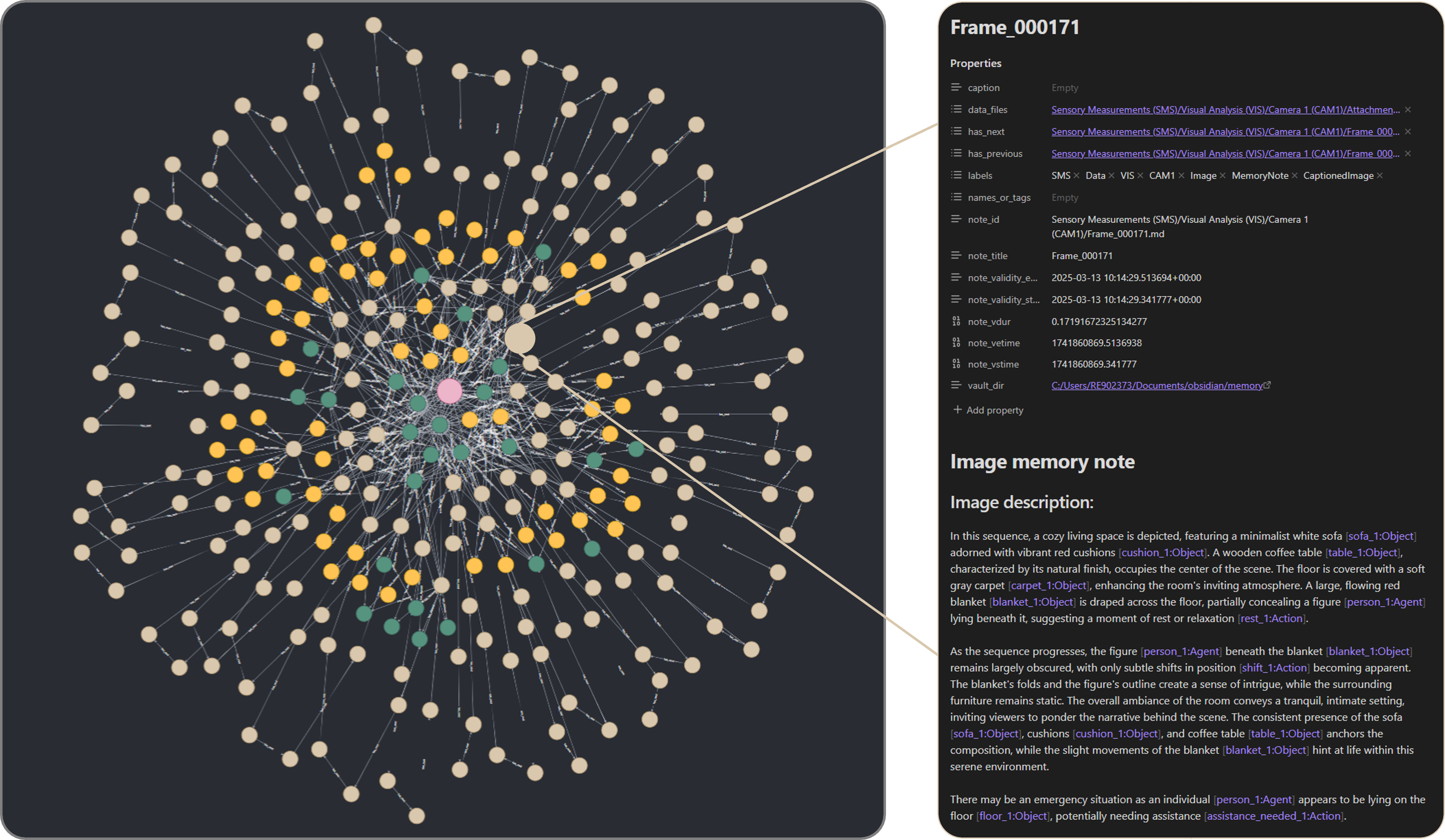}
  \caption{Knowledge graph (left) and image note (right). Sand-colored nodes represent sequential video frames, green nodes denote detected object instances, the pink node corresponds to the agent, and yellow nodes represent actions. Entities are linked across frames and via action patterns.}
  \label{fig:example-graph}
\end{figure}

The information captured during the perception phase, cp.~\Cref{subsec:perception}, is stored in a hybrid knowledge base combining a knowledge graph consisting of structured relationships and a vector store, i.e., a text-oriented representation allowing for semantic search.
The ingestion process consists of four steps.
First, all entities identified during the perception phase are extracted from the image captions.
Together with the entity names, we extract the entity types.
Second, we create embedding vectors for the image captions using an embedding model, resulting in high-dimensional numeric representations of the text.
If necessary due to context window limitations, the captions are split up into several parts.
Third, we create nodes in the knowledge graph for all images and connect them sequentially.
To these we add the respective attributes, such as the captions and the paths for the image files, and the embedding vectors for the image captions.
Fourth, we add nodes for all actions, agents, and objects identified and connect them to all the image nodes in which they appear, turning the sequence of images into a connected graph.
The knowledge graph, cp.~\Cref{fig:example-graph}, maintains temporal order via sequentially connected image nodes, while objects, agents, and actions structure events.
Consistent entity labels ensure continuity and enable context-aware retrieval.

%

\subsection{Agentic Retrieval for Question Answering}
\label{subsec:retrieval}





Combining a graph-based structured representation with natural language text notes allows the use of several retrieval techniques.
%
First, we use the memory system as a standard \gls{rag} system for \textit{semantic search} as it is based on natural language notes with embedding vectors associated to them.
By embedding the user query with the same embedding model, we create an embedding vector which we compare to the embeddings of all the notes, retrieving the semantically most relevant notes.
To increase the relevance of the context provided to the \gls{llm}, we optionally use a reranker to check the retrieved notes and filter out less relevant ones.
Providing the retrieved notes as context, we let the \gls{llm} answer the original question.
This type of retrieval is especially efficient for questions which are likely to have responses that are semantically close.
%
Second, we leverage the structure of the memory system for \textit{graph expansion}.
Relevant information may be included in notes which are not found when relying purely on semantic search, but that are linked to the notes found in the graph.
This is usually the case for relevant background information, e.g., personal preferences not showing up in individual notes, but represented in a note for an agent.
Here, we start with regular semantic search for identifying an initial set of relevant notes.
Then we expand the search results using an expansion algorithm.
Specifically, we use PageRank~\cite{rank1998pagerank}, but other algorithms, e.g., random walks, also work.
Finally, the expanded search results are used as context for \gls{llm}-based question answering.
This second type of retrieval is beneficial whenever there is implicit background information, which cannot easily be found via semantic search, but which becomes apparent when analyzing the surrounding environment of the relevant nodes in the graph.
%
Third, we rely on the graph representation for structured information retrieval in the form of \textit{text2cypher}.
For this, we let the \gls{llm} translate the user input into a Cypher query, which is run against the graph database.
The result in the form of a table is interpreted by the \gls{llm}, which formulates a natural language response.
This type of retrieval is ideal for answering structure-oriented queries, for instance questions that require counting entities.
%
To leverage the benefits of all three retrieval techniques, we combine them in an agentic retrieval system.
For this, we wrap the three retrieval functionalities into tools that we provide to an \gls{llm}-based agent, who can access them as needed for answering user questions.
The agent is prompted to select the most suitable tool, or several if necessary, to retrieve information from the memory system and eventually formulate an appropriate answer.

\begin{wrapfigure}{r}{0.5\textwidth}
  \centering
  \includegraphics[width=\linewidth]{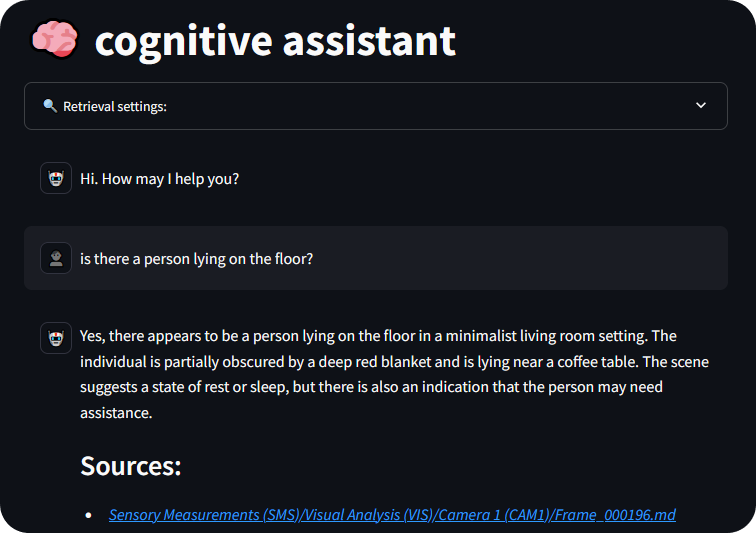}
  \caption{Chat frontend for interacting with the memory system.}
  \label{fig:frontend}
\end{wrapfigure}

\section{Implementation Details}
\label{sec:implementation-details}
We implemented a prototype using Neo4j as the graph database and vector store. Videos were recorded using an iPhone 15, split into frames at 3 Hz, and captions were generated for every fifth frame using \texttt{gpt-4o} for vision and \texttt{text-embedding-3-small} for storing embeddings in the vector store.
New entities (images, actions, agents, objects) are inserted into the graph using dynamically generated Cypher queries. Memory notes are also stored in Obsidian, enabling easy browsing. A minimalist gradio frontend allows users to interact with the system via a chat interface,~cp.~\Cref{fig:frontend}, that provides links to the sources used for transparency.
\Cref{tab:examples} presents representative questions and system responses, demonstrating retrieval capabilities across different query types.


\begin{table}[ht]
\begin{center}
    \begin{tabularx}{\linewidth}{ p{1.5cm} | p{3.2cm} | X }
        \textbf{Category} & \textbf{Question} & \textbf{System answer} \\
        \hline
        Counting in frame & How many images are there on the wall? & There are five photographs pinned to the wall, adding a personal touch to the minimalist living space. \\
        \hline
        Counting across frames & How many people are there? & There is 1 person present. \\
        \cline{2-3}
        & How many images are there in memory? & There are 329 images in memory. \\
        \hline
        \multirow{2}{*}{Existence} & Is there a person lying on the floor? & Yes, there is a person lying on the floor. \\
        \cline{2-3}
         & Is there something dangerous happening? & An individual appears to be lying on the floor, potentially needing assistance or indicating a medical emergency. \\
         \hline
         Emotional & What is the person's emotional state? & The person's emotional state appears to be relaxed and contemplative, as they are depicted reclining on a sofa, enveloped in a vibrant red blanket. \\
         \hline
         \multirow{2}{*}{Descriptive} & Please describe the setting of the scene. & The setting of the scene is a minimalist living space characterized by a light-colored wall and a soft gray carpet. Central to the scene is a modern white sofa adorned with vibrant red accent pillows, creating a cozy and inviting atmosphere. A wooden coffee table with a natural finish is positioned in front of the sofa, and it features a remote control, suggesting a casual and comfortable environment. \\
         \cline{2-3}
         & What is the man wearing? & The man is wearing a dark navy sweater and gray pants. He is also depicted with short, neatly styled hair and glasses, which add to his studious demeanor. \\
    \end{tabularx}
\end{center}
\caption{Representative examples for questions and system answers (all of which are correct).}
\label{tab:examples}
\end{table}

\section{Summary and Outlook}



This paper presents a grounded memory system that integrates the strengths of a knowledge graph and a vector store for agentic \gls{rag} with an \gls{llm} as an intuitive natural language interface
The system leverages a minimalist schema and operates through three key phases: perception, ingestion, and retrieval.
The system has potential applications ranging from robotics to assistive technologies, such as support systems for dementia patients.
Our approach provides a foundation for structured memory-based retrieval and serves as a starting point for future research in long-term knowledge representation and context-aware reasoning.
The integration of conceptual nodes provides additional flexibility, allowing retrieval to be guided by semantic relationships rather than purely temporal order.
This structured approach enables conversational assistants to reason over past events, improving long-term memory consistency compared to standard \gls{rag} techniques.



Future work should focus on scaling up the system and conducting large-scale evaluations in real-world scenarios.
Expanding to longer multimodal sequences will allow the system to capture broader temporal dependencies and leverage its effectiveness in retrieving and reasoning over complex event histories.
While we expect challenges in long-term entity disambiguation -- ensuring that agents, objects, and actions are consistently recognized across different scenes and timeframes -- moving beyond individual observations enables high-level behavior pattern identification by recognizing repetitive actions, activity trends, and structured sequences of human interactions. 
Additionally, we will further advance \gls{rag} techniques, such as recursive summarization and query rewriting, to enhance contextual understanding and improve response accuracy.
This can be supported by advancing agency in the retriever and providing further retrieval tools, e.g., via frameworks focusing on leveraging large sets of tools for \glspl{llm}~\cite{ocker2024tulip}.
Further improvements will also involve extending the system’s multimodal capabilities beyond vision, incorporating audio and spatial information.
In doing so, we aim to move toward a comprehensive memory system capable of supporting autonomous agents in complex environments.

\bibliography{sample-ceur}

\begin{thebibliography}{23}
\expandafter\ifx\csname natexlab\endcsname\relax\def\natexlab#1{#1}\fi
\providecommand{\url}[1]{\texttt{#1}}
\providecommand{\href}[2]{#2}
\providecommand{\path}[1]{#1}
\providecommand{\DOIprefix}{doi:}
\providecommand{\ArXivprefix}{arXiv:}
\providecommand{\URLprefix}{URL: }
\providecommand{\Pubmedprefix}{pmid:}
\providecommand{\doi}[1]{\href{http://dx.doi.org/#1}{\path{#1}}}
\providecommand{\Pubmed}[1]{\href{pmid:#1}{\path{#1}}}
\providecommand{\bibinfo}[2]{#2}
\ifx\xfnm\relax \def\xfnm[#1]{\unskip,\space#1}\fi
\bibitem[{Kant(1908)}]{kant1908critique}
\bibinfo{author}{I.~Kant},
\newblock \bibinfo{title}{Critique of pure reason. 1781},
\newblock \bibinfo{journal}{Modern Classical Philosophers, Cambridge, MA: Houghton Mifflin}  (\bibinfo{year}{1908}) \bibinfo{pages}{370--456}.
\bibitem[{Kashmira et~al.(2024)Kashmira, Dantanarayana, Brodsky, Mahendra, Kang, Flautner, Tang, and Mars}]{kashmira2024graph}
\bibinfo{author}{S.~Kashmira}, \bibinfo{author}{J.~L. Dantanarayana}, \bibinfo{author}{J.~Brodsky}, \bibinfo{author}{A.~Mahendra}, \bibinfo{author}{Y.~Kang}, \bibinfo{author}{K.~Flautner}, \bibinfo{author}{L.~Tang}, \bibinfo{author}{J.~Mars},
\newblock \bibinfo{title}{A graph-based approach for conversational {AI}-driven personal memory capture and retrieval in a real-world application},
\newblock \bibinfo{journal}{arXiv:2412.05447}  (\bibinfo{year}{2024}).
\bibitem[{Eggert et~al.(2020)Eggert, Deigm{\"o}ller, Fischer, and Richter}]{eggert2020action}
\bibinfo{author}{J.~Eggert}, \bibinfo{author}{J.~Deigm{\"o}ller}, \bibinfo{author}{L.~Fischer}, \bibinfo{author}{A.~Richter},
\newblock \bibinfo{title}{Action representation for intelligent agents using {Memory Nets}},
\newblock in: \bibinfo{booktitle}{IC3K}, \bibinfo{year}{2020}.
\bibitem[{Al-Tawil et~al.(2024)Al-Tawil, Hempel, Abdelrahman, and Al-Hamadi}]{al2024review}
\bibinfo{author}{B.~Al-Tawil}, \bibinfo{author}{T.~Hempel}, \bibinfo{author}{A.~Abdelrahman}, \bibinfo{author}{A.~Al-Hamadi},
\newblock \bibinfo{title}{A review of visual slam for robotics: Evolution, properties, and future applications},
\newblock \bibinfo{journal}{Frontiers in Robotics and AI} \bibinfo{volume}{11} (\bibinfo{year}{2024}) \bibinfo{pages}{1347985}.
\bibitem[{Armeni et~al.(2019)Armeni, He, Gwak, Zamir, Fischer, Malik, and Savarese}]{armeni20193d}
\bibinfo{author}{I.~Armeni}, \bibinfo{author}{Z.-Y. He}, \bibinfo{author}{J.~Gwak}, \bibinfo{author}{A.~R. Zamir}, \bibinfo{author}{M.~Fischer}, \bibinfo{author}{J.~Malik}, \bibinfo{author}{S.~Savarese},
\newblock \bibinfo{title}{{3D} scene graph: A structure for unified semantics, {3D} space, and camera},
\newblock in: \bibinfo{booktitle}{ICCV}, \bibinfo{year}{2019}.
\bibitem[{Kim et~al.(2019)Kim, Park, Song, and Kim}]{kim20193}
\bibinfo{author}{U.-H. Kim}, \bibinfo{author}{J.-M. Park}, \bibinfo{author}{T.-J. Song}, \bibinfo{author}{J.-H. Kim},
\newblock \bibinfo{title}{{3D} scene graph: A sparse and semantic representation of physical environments for intelligent agents},
\newblock \bibinfo{journal}{IEEE transactions on cybernetics} \bibinfo{volume}{50} (\bibinfo{year}{2019}) \bibinfo{pages}{4921--4933}.
\bibitem[{Rosinol et~al.(2020)Rosinol, Gupta, Abate, Shi, and Carlone}]{rosinol20023d}
\bibinfo{author}{A.~Rosinol}, \bibinfo{author}{A.~Gupta}, \bibinfo{author}{M.~Abate}, \bibinfo{author}{J.~Shi}, \bibinfo{author}{L.~Carlone},
\newblock \bibinfo{title}{{3D} dynamic scene graphs: Actionable spatial perception with places, objects, and humans},
\newblock in: \bibinfo{booktitle}{RSS}, \bibinfo{year}{2020}.
\bibitem[{Agia et~al.(2022)Agia, Jatavallabhula, Khodeir, Miksik, Vineet, Mukadam, Paull, and Shkurti}]{agia2022taskography}
\bibinfo{author}{C.~Agia}, \bibinfo{author}{K.~M. Jatavallabhula}, \bibinfo{author}{M.~Khodeir}, \bibinfo{author}{O.~Miksik}, \bibinfo{author}{V.~Vineet}, \bibinfo{author}{M.~Mukadam}, \bibinfo{author}{L.~Paull}, \bibinfo{author}{F.~Shkurti},
\newblock \bibinfo{title}{Taskography: Evaluating robot task planning over large {3D} scene graphs},
\newblock in: \bibinfo{booktitle}{CoRL}, \bibinfo{year}{2022}.
\bibitem[{Xie et~al.(2024)Xie, Min, Zhang, Xu, Bajaj, Salakhutdinov, Johnson-Roberson, and Bisk}]{xie2024embodied}
\bibinfo{author}{Q.~Xie}, \bibinfo{author}{S.~Y. Min}, \bibinfo{author}{T.~Zhang}, \bibinfo{author}{K.~Xu}, \bibinfo{author}{A.~Bajaj}, \bibinfo{author}{R.~Salakhutdinov}, \bibinfo{author}{M.~Johnson-Roberson}, \bibinfo{author}{Y.~Bisk},
\newblock \bibinfo{title}{Embodied-{RAG}: General non-parametric embodied memory for retrieval and generation},
\newblock \bibinfo{journal}{arXiv:2409.18313}  (\bibinfo{year}{2024}).
\bibitem[{Wang et~al.(2023)Wang, Yang, and Ren}]{wang2023lifelongmemory}
\bibinfo{author}{Y.~Wang}, \bibinfo{author}{Y.~Yang}, \bibinfo{author}{M.~Ren},
\newblock \bibinfo{title}{{LifelongMemory}: Leveraging {LLMs} for answering queries in long-form egocentric videos},
\newblock \bibinfo{journal}{arXiv:2312.05269}  (\bibinfo{year}{2023}).
\bibitem[{Fan et~al.(2024)Fan, Ma, Wu, Du, Li, Gao, and Li}]{fan2024videoagent}
\bibinfo{author}{Y.~Fan}, \bibinfo{author}{X.~Ma}, \bibinfo{author}{R.~Wu}, \bibinfo{author}{Y.~Du}, \bibinfo{author}{J.~Li}, \bibinfo{author}{Z.~Gao}, \bibinfo{author}{Q.~Li},
\newblock \bibinfo{title}{{VideoAgent}: A memory-augmented multimodal agent for video understanding},
\newblock in: \bibinfo{booktitle}{ECCV}, \bibinfo{year}{2024}.
\bibitem[{Goletto et~al.(2024)Goletto, Nagarajan, Averta, and Damen}]{goletto2024amego}
\bibinfo{author}{G.~Goletto}, \bibinfo{author}{T.~Nagarajan}, \bibinfo{author}{G.~Averta}, \bibinfo{author}{D.~Damen},
\newblock \bibinfo{title}{Amego: Active memory from long egocentric videos},
\newblock in: \bibinfo{booktitle}{ECCV}, \bibinfo{year}{2024}.
\bibitem[{Li et~al.(2025)Li, Lai, Bao, Tan, Dao, Sui, Shen, Liu, Liu, and Kong}]{li2025visual}
\bibinfo{author}{Y.~Li}, \bibinfo{author}{Z.~Lai}, \bibinfo{author}{W.~Bao}, \bibinfo{author}{Z.~Tan}, \bibinfo{author}{A.~Dao}, \bibinfo{author}{K.~Sui}, \bibinfo{author}{J.~Shen}, \bibinfo{author}{D.~Liu}, \bibinfo{author}{H.~Liu}, \bibinfo{author}{Y.~Kong},
\newblock \bibinfo{title}{Visual large language models for generalized and specialized applications},
\newblock \bibinfo{journal}{arXiv:2501.02765}  (\bibinfo{year}{2025}).
\bibitem[{Hogan et~al.(2021)Hogan, Blomqvist, Cochez, d’Amato, Melo, Gutierrez, Kirrane, Gayo, Navigli, Neumaier et~al.}]{hogan2021knowledge}
\bibinfo{author}{A.~Hogan}, \bibinfo{author}{E.~Blomqvist}, \bibinfo{author}{M.~Cochez}, \bibinfo{author}{C.~d’Amato}, \bibinfo{author}{G.~D. Melo}, \bibinfo{author}{C.~Gutierrez}, \bibinfo{author}{S.~Kirrane}, \bibinfo{author}{J.~E.~L. Gayo}, \bibinfo{author}{R.~Navigli}, \bibinfo{author}{S.~Neumaier}, et~al.,
\newblock \bibinfo{title}{Knowledge graphs},
\newblock \bibinfo{journal}{ACM Computing Surveys} \bibinfo{volume}{54} (\bibinfo{year}{2021}) \bibinfo{pages}{1--37}.
\bibitem[{Eggert and Ocker(2025)}]{eggert2025graph}
\bibinfo{author}{J.~Eggert}, \bibinfo{author}{F.~Ocker}, \bibinfo{title}{Graph based memory extension for large language models}, \bibinfo{year}{2025}. \bibinfo{note}{US Patent App. 18/898,607}.
\bibitem[{Maniar et~al.(2025)Maniar, Chan, Zulfikar, Ren, Xu, and Maes}]{maniar2025mempal}
\bibinfo{author}{N.~Maniar}, \bibinfo{author}{S.~W. Chan}, \bibinfo{author}{W.~Zulfikar}, \bibinfo{author}{S.~Ren}, \bibinfo{author}{C.~Xu}, \bibinfo{author}{P.~Maes},
\newblock \bibinfo{title}{{MemPal}: Leveraging multimodal {AI} and {LLMs} for voice-activated object retrieval in homes of older adults},
\newblock in: \bibinfo{booktitle}{IUI}, \bibinfo{year}{2025}.
\bibitem[{Li et~al.(2024)Li, Zhang, and Ma}]{li2024omniquery}
\bibinfo{author}{J.~N. Li}, \bibinfo{author}{Z.~J. Zhang}, \bibinfo{author}{J.~Ma},
\newblock \bibinfo{title}{Omniquery: Contextually augmenting captured multimodal memory to enable personal question answering},
\newblock \bibinfo{journal}{arXiv:2409.08250}  (\bibinfo{year}{2024}).
\bibitem[{Guti{\'e}rrez et~al.(2024)Guti{\'e}rrez, Shu, Gu, Yasunaga, and Su}]{gutierrez2024hipporag}
\bibinfo{author}{B.~J. Guti{\'e}rrez}, \bibinfo{author}{Y.~Shu}, \bibinfo{author}{Y.~Gu}, \bibinfo{author}{M.~Yasunaga}, \bibinfo{author}{Y.~Su},
\newblock \bibinfo{title}{{HippoRAG}: Neurobiologically inspired long-term memory for large language models},
\newblock in: \bibinfo{booktitle}{NeurIPS}, \bibinfo{year}{2024}.
\bibitem[{Ozsoy et~al.(2024)Ozsoy, Messallem, Besga, and Minneci}]{ozsoy2024text2cypher}
\bibinfo{author}{M.~G. Ozsoy}, \bibinfo{author}{L.~Messallem}, \bibinfo{author}{J.~Besga}, \bibinfo{author}{G.~Minneci},
\newblock \bibinfo{title}{Text2cypher: Bridging natural language and graph databases},
\newblock \bibinfo{journal}{arXiv:2412.10064}  (\bibinfo{year}{2024}).
\bibitem[{Edge et~al.(2024)Edge, Trinh, Cheng, Bradley, Chao, Mody, Truitt, and Larson}]{edge2024local}
\bibinfo{author}{D.~Edge}, \bibinfo{author}{H.~Trinh}, \bibinfo{author}{N.~Cheng}, \bibinfo{author}{J.~Bradley}, \bibinfo{author}{A.~Chao}, \bibinfo{author}{A.~Mody}, \bibinfo{author}{S.~Truitt}, \bibinfo{author}{J.~Larson},
\newblock \bibinfo{title}{From local to global: A graph {RAG} approach to query-focused summarization},
\newblock \bibinfo{journal}{arXiv:2404.16130}  (\bibinfo{year}{2024}).
\bibitem[{Achiam et~al.(2024)}]{achiam2024gpt4}
\bibinfo{author}{J.~Achiam}, et~al.,
\newblock \bibinfo{title}{Gpt-4 technical report},
\newblock \bibinfo{journal}{arXiv:2303.08774}  (\bibinfo{year}{2024}).
\bibitem[{Bin and Page(1998)}]{rank1998pagerank}
\bibinfo{author}{S.~Bin}, \bibinfo{author}{K.~L. Page},
\newblock \bibinfo{title}{The anatomy of a large-scale hypertextual web search engine},
\newblock in: \bibinfo{booktitle}{Computer Networks}, \bibinfo{year}{1998}.
\bibitem[{Ocker et~al.(2024)Ocker, Tanneberg, Eggert, and Gienger}]{ocker2024tulip}
\bibinfo{author}{F.~Ocker}, \bibinfo{author}{D.~Tanneberg}, \bibinfo{author}{J.~Eggert}, \bibinfo{author}{M.~Gienger},
\newblock \bibinfo{title}{Tulip agent--enabling {LLM}-based agents to solve tasks using large tool libraries},
\newblock \bibinfo{journal}{arXiv preprint arXiv:2407.21778}  (\bibinfo{year}{2024}).

\end{thebibliography}


\end{document}